\pdfoutput=1

\documentclass[11pt]{article}

\usepackage[]{acl}

\usepackage{times}
\usepackage{latexsym}
\usepackage{paralist}
\usepackage[most]{tcolorbox}

\usepackage[T1]{fontenc}

\usepackage[utf8]{inputenc}

\usepackage{microtype}

\usepackage{inconsolata}

\usepackage{graphicx}
\usepackage{array}
\usepackage{multicol}
\usepackage{tabularray}
\usepackage{tabularx}
\usepackage{booktabs, multirow} 
\usepackage{soul}
\usepackage{tabularx}

\title{Visualizing Dialogues: Enhancing Image Selection through\\ Dialogue Understanding with Large Language Models
}

\author{Chang-Sheng Kao \and Yun-Nung Chen \\
    National Taiwan University, Taipei, Taiwan \\
    \texttt{r11922a14@csie.ntu.edu.tw} \and \texttt{y.v.chen@ieee.org} \\}

\begin{document}
\maketitle

\begin{abstract}
Recent advancements in dialogue systems have highlighted the significance of integrating multimodal responses, which enable conveying ideas through diverse modalities rather than solely relying on text-based interactions. This enrichment not only improves overall communicative efficacy but also enhances the quality of conversational experiences. However, existing methods for dialogue-to-image retrieval face limitations due to the constraints of pre-trained vision language models (VLMs) in comprehending complex dialogues accurately. To address this, we present a novel approach leveraging the robust reasoning capabilities of large language models (LLMs) to generate precise dialogue-associated visual descriptors, facilitating seamless connection with images. Extensive experiments conducted on benchmark data validate the effectiveness of our proposed approach in deriving concise and accurate visual descriptors, leading to significant enhancements in dialogue-to-image retrieval performance. Furthermore, our findings demonstrate the method's generalizability across diverse visual cues, various LLMs, and different datasets, underscoring its practicality and potential impact in real-world applications.\footnote{\url{https://github.com/MiuLab/VisualDialog}}

\end{abstract}

\section{Introduction}

In recent years, the landscape of online conversations has undergone a significant transformation thanks to the proliferation of instant messaging tools.
Unlike the past, when these exchanges were confined to text alone, today's conversations have evolved into a multimodal experience, incorporating elements like images and speech.
The various communication modes not only enhance engagement but also prove invaluable for conveying complex information that can be challenging to communicate solely through text. 
\citet{sun-etal-2022-multimodal} highlighted the advantages of integrating images into conversations.
For example, when discussing a topic with someone who may not grasp the concept, sharing an image can provide visual clarity for better comprehension.
Additionally, when precision is required to convey specific details about a subject, relevant images can be a more effective means of communication than text alone~\cite{hsu2023visually}. 
Consequently, the ability to generate responses using images is a crucial area of research in enhancing automatic dialogue systems. 
To equip these systems with the capacity to respond using images, a common method involves text-to-image retrieval, as demonstrated by previous work \cite{liao2018knowledge,zang-etal-2021-photochat}.
In this approach, a model selects an appropriate image from a pre-constructed image repository based on the context of the ongoing conversation.


As storage costs decline and computational power advances, vision foundation models pre-trained on large-scale, open-domain image-text pairs have emerged \cite{radford2021learning,jia2021scaling,yuan2021florence}. These models have demonstrated outstanding performance in text-to-image retrieval tasks, excelling in both zero-shot and fully-trained scenarios. 
However, despite their impressive capabilities, these pre-trained vision-language models (VLMs) still come with some limitations.
One significant limitation is their suboptimal design for handling complete dialogue contexts effectively. 
Often, they suffer from extracting key information comprehensively from the entire conversation. 
Table~\ref{tab:example} presents an illustrative example, where a dialogue-to-image model fine-tuned from CLIP~\cite{radford2021learning} fails to correctly interpret the dialogue's intent. 
This highlights the challenge of dialogue comprehension, a task for which pre-trained VLMs may not be adequately equipped. 
Additionally, most existing VLMs typically impose input text length constraints during their pre-training stages, preventing them from processing the entirety of the dialogue context directly.
This constraint can lead to the loss of crucial contextual information, potentially undermining the model's overall performance.

\begin{table}[t!]\centering
\small
\begin{tabular}{p{70mm}}\toprule
\bf Dialogue context \\
\textbf{B:} how are you doing? \\
\textbf{A:} I'm doing good. Just out at a restaurant taking pictures for customers. \\
\textbf{B:} congratulations \\
\textbf{A:} It's hilarious watching people try to use chopsticks \\
\textbf{B:} i'm really happy for you friend \\
\textbf{B:} yeah, its really funny \\
\textbf{A:} Yeah, it's better than most gigs I get \\
\textbf{B:} even i still try to try to find a way around that thing \\
\textbf{A:} I give up and ask for a fork. I want that rice in my mouth!!!!! \\
\textbf{A:} \colorbox{yellow}{(share a photo)} \\
\end{tabular}

\begin{tabular}{p{32.5mm}p{32.5mm}}\toprule
\bf Ground-truth & \bf Retrieved top-1\\
\includegraphics[width=0.2\textwidth,height=0.2\textwidth]{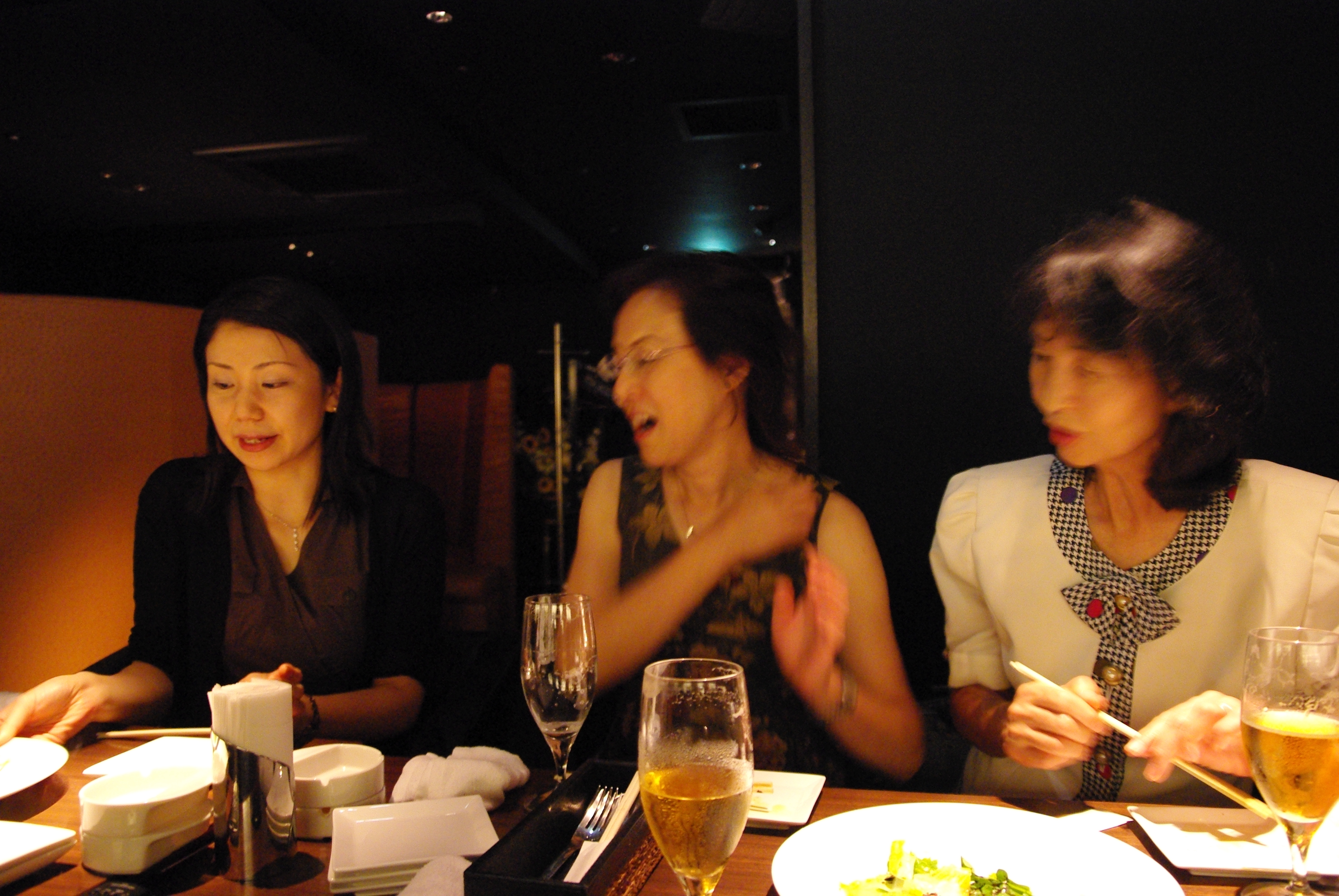}  &
\includegraphics[width=0.2\textwidth,height=0.2\textwidth]{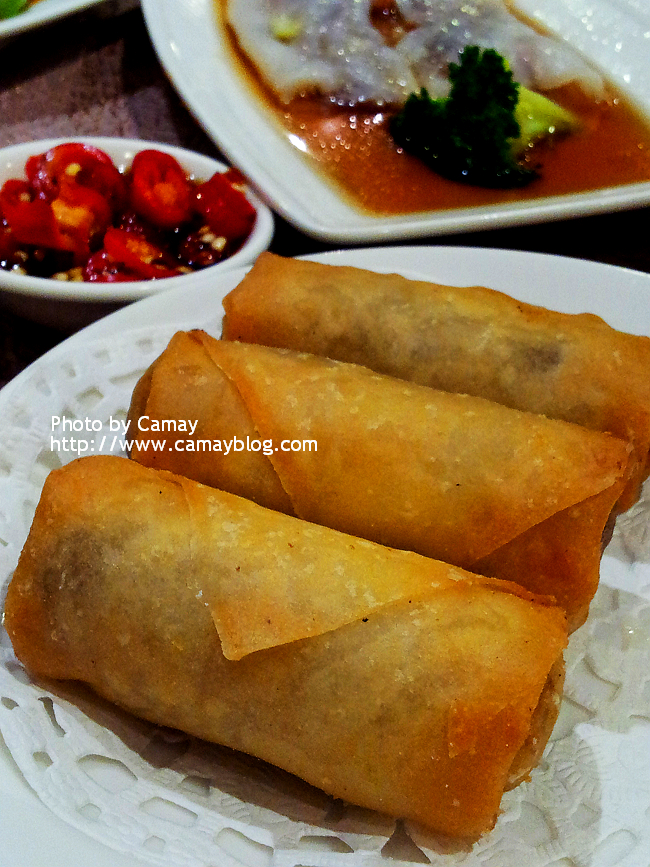}\\
\midrule
\multicolumn{2}{l}{\bf Dialogue-associated visual cues}\\
\multicolumn{2}{l}{\labelitemi\hspace{\dimexpr\labelsep+0.5\tabcolsep}main subject: \textcolor{red}{customers}} \\
\multicolumn{2}{l}{\labelitemi\hspace{\dimexpr\labelsep+0.5\tabcolsep}foreground objects: \textcolor{red}{chopsticks}, \textcolor{blue}{table}, \textcolor{blue}{food}} \\
\multicolumn{2}{l}{\labelitemi\hspace{\dimexpr\labelsep+0.5\tabcolsep}background scene: \textcolor{orange}{restaurant}} \\
\multicolumn{2}{l}{\labelitemi\hspace{\dimexpr\labelsep+0.5\tabcolsep}events: \textcolor{orange}{eating food}} \\
\bottomrule
\end{tabular} 
\vspace{-2mm}
\caption{An example of a dialogue and the shared image; the fine-tuned CLIP model fails to retrieve the correct one. \textcolor{red}{Red} indicates the missing elements, \textcolor{blue}{blue} indicates a perfect match, and \textcolor{orange}{orange} suggests a partial match.}
\label{tab:example}
\vspace{-2mm}
\end{table}

Inspired by \citet{menon2022visual}, we leverage the reasoning capabilities of large language models (LLMs) to generate the visual descriptor for the dialogue context.
These descriptors encapsulate speculations about the image that the speaker intends to share, aiming to provide concise and precise cues for better text-to-image retrieval. 
Our objective is to address the aforementioned limitations and enhance task performance.
Given that most vision models excel at identifying objects, scenes, and other visual elements in images~\cite{kuznetsova2020open,zang-etal-2021-photochat}, we employ a set of visually-focused queries, such as \textit{main subject} and \textit{background scene}, to bridge the gap between the ongoing dialogue and the pool of potential image candidates.
These queries serve as templates for the LLM to predict corresponding visual cues based on the dialogue context.
We then utilize these queries and their resulting answers as dialogue-associated visual descriptors, as illustrated in the bottom part of Table~\ref{tab:example}. 
Our experiments on the benchmark dataset showcase the exceptional performance of our approaches, surpassing all previous results.
In addition to demonstrating the effectiveness of our LLM-generated visual descriptor, we compare it with other descriptor creation methods and conduct an in-depth analysis to evaluate the efficacy of each proposed query.

Our contributions can be summarized as 3-fold:
\begin{compactitem}
    \item This paper introduces a novel approach for retrieving associated photos in dialogue systems, leveraging the reasoning capabilities of LLMs to generate visually-focused cues for improved image retrieval.
    \item We design a series of visually-focused queries based on common image features, employing them to construct conversation descriptors. 
    Our experiments validate the effectiveness of these designed queries.
    \item The proposed approach achieves state-of-the-art performance on multiple benchmark datasets.
\end{compactitem}

\section{Related Work}

\begin{figure*}[t]
\centering
\includegraphics[width=.9\textwidth]{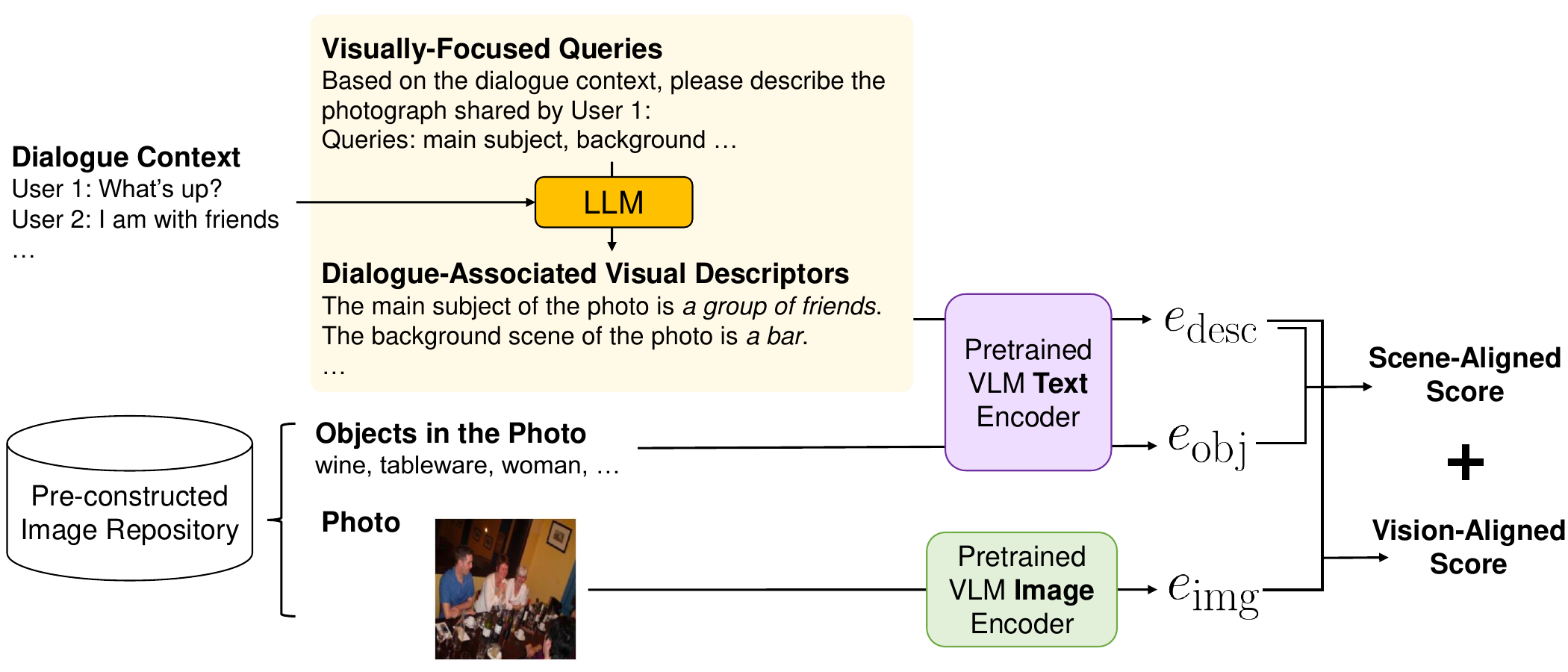}
\vspace{-2mm}
\caption{
The framework of our proposed method. We employ the text encoder from a pre-trained VLM to encode both the descriptor and the object list. This yields two distinctive features, namely the descriptor embedding ($e_\text{desc}$) and the object list feature ($e_\text{obj}$). Additionally, we utilize the VLM's image encoder to process and encode the image, resulting in the image embedding ($e_\text{img}$). The final retrieval score is then computed by aggregating a scene-aligned score and a vision-aligned score.}
\label{img:framework}
\vspace{-2mm}
\end{figure*}

\paragraph{Multimodal Dialogue Systems}

Recent years have witnessed a notable shift in research towards multimodal dialogues, moving beyond the confines of text-only interactions \citep{9948337}. 
While the exploration of image-grounded conversations, where textual dialogues are generated from images, has gained traction \cite{yang2021open,shuster-etal-2021-multi,chen2023multimodal,gong2023multimodalgpt,zhu2023minigpt,liu2023llava}, an increasing number of studies are delving into the incorporation of multimodal responses within dialogue systems.
This multimodal evolution enables human-machine conversations to reflect real-life human-human interactions and communicate concepts that are difficult to convey through text alone. 
For instance, \citet{liao2018knowledge} introduced a task-oriented multimodal dialogue system featuring a taxonomy-based learning module that captures nuanced visual semantics and employs reinforcement learning to ensure response coherence.
Further advancements in robotic technology now enable robots to more effectively understand and respond to human requests in life support scenarios. This requires the interpretation of both the visual context of the environment and the user’s verbal communication, integrating multimodal capabilities into dialogue scenarios~\cite{tanaka2024demand,tsai2024asmr}.
Moreover, \citet{sun-etal-2022-multimodal} introduced a framework capable of directly generating multimodal responses via a text-to-target-modality generator. 
In contrast, rather than directly generating multimodal responses, \citet{zang-etal-2021-photochat} achieve multimodal responses by employing image retrieval models to select appropriate images from a pre-existing image repository.
For better practicality, our paper centers on the same task---recommending a suitable image from the user's image repository based on the ongoing dialogue context.

\paragraph{External Knowledge of LLMs for Visual Tasks}

Many studies have showcased that the commonsense knowledge and reasoning capabilities in large language models (LLMs) can significantly augment the performance of visual tasks.
For instance, \citet{NEURIPS2021_01b7575c} confirmed that by projecting image encodings into the embedding space of an LLM, it becomes possible to harness the rich knowledge contained within the LLM for few-shot visual question answering (VQA) tasks.
Similarly, \citet{zeng2022socraticmodels} introduced Socratic Models, which leverages multiple pre-trained large models trained on data from diverse domains. 
By translating non-language domain information into textual prompts, Socratic Models achieve state-of-the-art results in zero-shot image captioning and video-to-text retrieval tasks.
Furthermore, \citet{menon2022visual} took a novel approach by obtaining visual features for different categories through queries to GPT-3 \citep{NEURIPS2020_1457c0d6} using category names.
These textual descriptors are then employed as internal representations for zero-shot visual classification and text-to-image retrieval tasks.
Our work centers on harnessing the reasoning capabilities of LLMs to derive contextually relevant visual descriptions for shared photos within the dialogue context.
Different from the prior work based on non-language domains or single sentences, our approach focuses on the nuanced domain of photo sharing within conversations, which presents unique challenges due to its reliance on commonsense knowledge and an understanding of human-human interactions.

\section{Methodology}

Our objective is to select an image from a pre-constructed photo set $\{(v_j, o_j)\}_{j=1}^m$ given a dialogue context $D$, where $v_j$ represents an image candidate and $o_j$ lists the objects appearing in $v_j$.
Note that the object lists can be obtained through object detection in the pre-processing stage, and we treat this object information as given data.

Figure~\ref{img:framework} illustrates the proposed framework, which introduces an innovative approach to estimate retrieval scores for each image candidate within a dialogue context.
These scores are based on two criteria: \textbf{scene-aligned} and \textbf{vision-aligned} scores, both relying on visual descriptors. 
The scene-aligned score assesses whether the speculated visual cues align with the image-associated objects in a \emph{textual} format. 
In contrast, the vision-aligned score evaluates the alignment between the visual description and the image using \emph{vision-language} models.



\begin{table*}[t]
\centering\small
\begin{tabular}{p{92mm} p{60mm}}
\toprule
\bf Visual Features \& Descriptions & \bf Examples\\
\midrule
\textbf{Main subject}: the photo-focused objects for conveying a particular theme & \it people, cakes, buildings\\ 
\textbf{Prominent objects in the foreground}: objects in addition to the main subject convey signal for photo understanding. & \it a bar counter and bottles in a photo taken at a bar\\ 
\textbf{Background scene}: the background scene in the photo & \it restaurants, bars, outdoors\\ 
\textbf{Events}: activities or events currently captured in the photo & \it weddings, birthdays, eating food\\
\textbf{Materials and attributes}: finer details about the photo & \it teapot made of ceramic, black and white feathers\\
\bottomrule
\end{tabular}
\vspace{-1mm}
\caption{\label{tab:descriptors} All designed descriptors used in the proposed method.}
\vspace{-2mm}
\end{table*}

\subsection{Dialogue-Associated Visual Descriptor}
\label{sec:dialogue-associated-visual-descriptor}

Considering that visual descriptors can significantly enhance the understanding of visual content~\cite{menon2022visual}, we focus on generating dialogue-associated visual descriptors to improve image retrieval capabilities. 
To create high-quality visual descriptors that can connect with visual elements in the photo, we define a set of visually-focused queries, denoted as $Q = \{q_i\}$. 
These queries encompass various visual attributes related to an image, such as \textit{main subject} and \textit{background scene}, which are instrumental in linking the target photo to the dialogue.


Drawing from prior work~\cite{kuznetsova2020open,zang-etal-2021-photochat} and our common experiences, we assume that photos shared in online messaging typically contain components such as \emph{main subjects}, \emph{prominent foreground objects}, \emph{background scenes}, \emph{events}, and \emph{materials and attributes}, as detailed in Table~\ref{tab:descriptors}.
Note that we do not expect all answers to these queries to be perfectly extracted from the dialogue context or found in the ground-truth image. 
Instead, our goal is to leverage automatically inferred visual descriptors to bridge the gap between the image and the given dialogue context.

Leveraging the powerful reasoning capabilities of large language models (LLMs) \cite{touvron2023llama}, we construct a prompt comprising the dialogue $D$ and the set of queries $Q$ and input it into the LLM.
This process yields a set of dialogue-associated visual descriptors in a zero-shot manner:
\begin{equation} \label{eq:LLM}
    \text{desc} = \text{LLM}(D, Q).
\end{equation}For instance, a generated visual descriptor regarding the main subject might read, ``\emph{The main subject of the photo is a group of friends}.''
The used prompts can be found in Appendix~\ref{appendix:prompts}.

\subsection{Image Relevance Estimation}

To measure the relevance of each image candidate in the context of a given dialogue $D$, we calculate two retrieval scores based on their generated visual descriptors $\text{desc}$: $S_\text{scene}(o_j, \text{desc})$ and $S_\text{vision}(v_j, \text{desc})$. 
The former score assesses if the objects in the photo candidate align with the inferred visual descriptors in their \emph{text-only} forms, referred to as the \textbf{scene-aligned} score.
The latter score evaluates if the photo candidate matches the visual descriptions through \emph{multimodal} methods, termed the \textbf{vision-aligned} score.


\subsection{Image Retrieval Learning}
Our task involves retrieving the target image from a pre-constructed photo set, and it can be approached in two settings:
1) zero-shot and 2) training with contrastive leanrning.


\subsubsection{Zero-Shot}

Using the descriptor $\text{desc}$ derived from the dialogue context $D$, we employ a pre-trained vision language model (VLM) for zero-shot text-to-image retrieval. 
This process yields two scores through its text encoder and image encoder, as illustrated in Figure~\ref{img:framework}. 
The final retrieval score is calculated as:
\begin{equation}\label{eq:retrieval score}
   S_\text{scene}(o_j, \text{desc}) + \lambda \cdot S_\text{vision}(v_j, \text{desc}),
\end{equation}
where $\lambda$ is a weighting parameter. The image with the highest score is selected in a zero-shot manner.

\subsubsection{Contrastive Learning}
To further enhance retrieval performance, we fine-tune the VLM model using the training set.
Following the pre-training stage outlined by \citet{radford2021learning}, we apply contrastive learning to optimize our dialogue-image retriever. 
During training, we randomly sample a minibatch of dialogue-associated descriptors and photo pairs, designating $(\text{desc}, {v^*, o^*})$ as the positive example, while the remaining $(b - 1)$ examples within the minibatch serve as negative examples. 
The contrastive losses are calculated separately for the scene and vision components, focusing on aligning dialogue-associated visual descriptors and the target photo.
\begin{eqnarray} 
    \mathcal{L}_\text{scene} = - \log \frac{\exp(S_\text{scene}(o^*, \text{desc}) / \tau)}{\sum_{j \in b} \exp(S_\text{scene}(o_j, \text{desc}) / \tau)},\nonumber\\
    \mathcal{L}_\text{vision} = - \log \frac{\exp(S_\text{scene}(v^*, \text{desc}) / \tau)}{\sum_{j \in b} \exp(S_\text{scene}(v_j, \text{desc}) / \tau)},\nonumber
\end{eqnarray}
where $\tau$ is the trainable temperature parameter.
The final training loss is a combination of these contrastive losses:
\begin{equation}\label{eq:total loss}
    \mathcal{L} = \frac{1}{b} \sum_{j \in b} ( \mathcal{L}_\text{scene} + \lambda\cdot \mathcal{L}_\text{vision} ),
\end{equation}
where $\lambda$ is a weighting parameter.
This approach optimizes our dialogue-image retrieval model through contrastive learning.

\begin{table*}[ht]
\centering
\resizebox{0.98\textwidth}{!}{
\begin{tabular}{|ll|c c c c|c c c c|}
\hline
\multirow{2}{*}{\bf Method} & \multirow{2}{*}{\bf LLM} & \multicolumn{4}{c|}{\bf Zero-Shot} & \multicolumn{4}{c|}{\bf Fully-Trained}\\
 &  & \textbf{R@1} & \textbf{R@5} & \textbf{R@10} & \bf Avg & \textbf{R@1} & \textbf{R@5} & \textbf{R@10} & \bf Avg\\
\hline\hline
VSE++$^\dag$ & - & - & - & - & - & 10.20 & 25.40 & 34.20 & 23.27 \\
SCAN$^\dag$ & - & - &  - &- & - & 10.40 & 27.00 & 37.10 & 24.83 \\
DE - Diag (BERT) & - & - & - & - & - & 12.88 & 35.13 & 47.75 & 31.92\\
DE - Diag (CLIP) & - & - & - & - & - & 14.76 & 35.78 & 47.12 & 32.55 \\
\hline
Desc - Diag & - & 16.00 & 30.90 & 37.70 & 28.20 & 40.35 & 58.77 & 66.88 & 55.33\\
Desc - Caption & BLIP-2 & - & - & - & - & 16.68 & 35.34 & 45.17 & 32.40\\
Desc - Summary & LLaMA-2-7B-Chat & 22.90 & 40.10 & 47.60 & 36.87 & 42.81 & 62.42 & 71.35 & 58.86 \\
Desc - Summary & LLaMA-2-13B-Chat & 24.40 & 40.50 & 48.30 & 37.73 & \bf 44.17 & 64.23 & 72.66 & 60.35 \\
Desc - Guessing & LLaMA-2-7B-Chat & \underline{27.60} & \underline{47.80} & \underline{58.10} & \underline{44.50} & 42.55 & 64.22 & 72.29 & 59.69 \\
Desc - Guessing & LLaMA-2-13B-Chat & \textbf{29.30} & \textbf{51.30} & \textbf{59.80} & \bf 46.80 & 43.18 & \bf 65.45 & \underline{73.43} & \underline{60.69}\\
Desc - Queries & LLaMA-2-7B-Chat & 22.60 & 42.20 & 50.40 & 38.40 & 37.34 & 57.52 & 66.62 & 53.83\\
Desc - Queries & LLaMA-2-13B-Chat & 26.40 & 45.80 & 55.10 & 42.43 & \underline{44.00} & \underline{64.78} & \bf 73.95 & \bf 60.91\\
\hline
\end{tabular}}
\vspace{-2mm}
\caption{\label{tab:all-result}
Retrieval performance for zero-shot and fully-trained settings (\%).
We employ the LLM with greedy decoding to ensure the correct format and reasoning capability.
Each number is the average over 10 runs with different random seeds.
\dag denotes that we directly report the numbers from \citet{zang-etal-2021-photochat}.
}
\vspace{-2mm}
\end{table*}

\section{Experiments}\label{sec: experiments}


For the LLM in (\ref{eq:LLM}), we utilized well-established LLMs with instruction tuning and reinforcement learning from human feedback (RLHF), including LLaMA-2-7B-Chat and LLaMA-2-13B-Chat \citep{touvron2023llama}. 
We employed greedy decoding for generating descriptors to ensure the correct format and reasoning capability.
Our pre-trained vision-language model (VLM) backbone is CLIP ViT-B/32, and VLM training is executed on a single NVIDIA GeForce RTX 2080 Ti GPU with a batch size of 56. 
We utilize the ADAM optimizer with an initial learning rate of 1e-5. 
The weighting parameter $\lambda$ was set to 1 to strike a balance between scene-alignment and vision-alignment.

\subsection{Data \& Metrics}\label{subsec: datasets}

Our approach is evaluated on a benchmark multimodal dialogue dataset called PhotoChat.
PhotoChat dataset \cite{zang-etal-2021-photochat} is characterized by open-domain, high-quality multimodal dialogues and comprises 10,917 images paired with 12,286 dialogues. Specifically, the dataset is divided into 10,286 instances for training, 1,000 for validation, and another 1,000 for testing. Each image in the dataset is accompanied by an associated object list presented in textual form. In each data instance, one photo is shared within the context of the conversation.

Given that this task can be formulated as an image retrieval task, we employed $Recall@k$ ($R@k$) as our evaluation metric.
During the training phase, we select the final model based on the highest $avg(R@1, R@5, R@10)$ score on the validation set. In the testing phase, for each dialogue instance, the trained models retrieved images from the candidate photos in the testing set.

\subsection{Baselines}\label{subsec: baseline}

We compare our approach against several established baselines:
\begin{compactitem}
  \item VSE++: \citet{faghri2018vse++} incorporated hard negatives in the ranking loss to learn visual-semantic embeddings for text-image retrieval.
  \item SCAN: \citet{lee2018stacked} utilized stacked cross attention to align image regions and words in a sentence and calculate image-text similarity. 
  \item Dual Encoder (DE): Previous work \cite{parekh-etal-2021-crisscrossed,zang-etal-2021-photochat} employed a dual encoder architecture, where one encoder processes the image and its object list using CLIP ViT-B/32 for images and FFNN for object features. For the dialogue encoder, two different text encoders were experimented with: CLIP ViT-B/32 Text and BERT \cite{devlin-etal-2019-bert} with an additional projection to ensure consistent dimensions. The retrieval similarity between the image and dialogue encodings is measured using dot product.
  
\end{compactitem}

\subsection{Descriptor Variants}
In addition to the query-based descriptors, we conduct experiments using the following descriptor variants for in-depth analysis:
\begin{compactitem}
\item \textbf{Desc - Diag} (whole dialogue as descriptors): All utterances are concatenated to form the descriptors, allowing the image retriever to utilize complete cues within the dialogue.
\item \textbf{Desc - Caption} (caption as descriptors): Inspired by \citet{li2023blip}, we performed zero-shot image captioning on images in the training set using BLIP-2. We then trained a text generator to create image captions as descriptors based on a given dialogue.
\item \textbf{Desc - Summary} (summary as descriptors): Descriptors are generated by LLMs based on a dialogue summary, offering a more concise representation of the conversation.
\item \textbf{Desc - Guessing} (visually-focused guessing as descriptors): LLMs are allowed to speculate about the features of the upcoming shared photo from the dialogue without being constrained by a specific query.
\item \textbf{Desc - Queries} (visually-focused query descriptors): Utilizing our designed visually-focused attributes as descriptors.
\end{compactitem}

\subsection{Results}
Table~\ref{tab:all-result} provides a comprehensive overview of the results for both zero-shot and fully-trained settings on PhotoChat dataset.
In zero-shot scenarios, \textbf{Desc - Guessing} emerges as the top-performing method among all results. 
Notably, \textbf{Desc - Queries} outperforms \textbf{Desc - Summary}, indicating that visually-focused queries and guessing contribute valuable information for linking the desired images.
We hypothesize that the poor performance of \textbf{Desc - Summary} is because the LLM summarize all content of the dialogue, rather than focusing solely on the visual information. Consequently, \textbf{Desc - Summary} often includes sentences that do not aid in text-to-image retrieval. As for why \textbf{Desc - Queries} did not perform better than \textbf{Desc - Guessing}, we hypothesize that it is because we construct our descriptors using the templates from \ref{sec:dialogue-associated-visual-descriptor}. These sentences tend to be less fluent and are not the types of sentences CLIP was familiar with during pre-training stage. In contrast, \textbf{Desc - Guessing} generates more fluent descriptors without any constraints, making them closer to the sentences seen by CLIP during pre-training stage. This gives \textbf{Desc - Guessing} an advantage in zero-shot scenarios.

In the fully-trained setting, the descriptor-based results (\textbf{Desc - Summary}, \textbf{Desc - Guessing}, \textbf{Desc - Queries}) with LLaMA-2-13B-Chat exhibit similar performance, with \textbf{Desc - Queries} achieving the highest average performance. 
These results validate the effectiveness of our proposed approach, demonstrating that the generated visual descriptors successfully facilitate the connection between associated images through the LLM's understanding of dialogue.
Additionally, it is evident that LLaMA-2-13B-Chat outperforms LLaMA-2-7B-Chat due to its stronger reasoning abilities for understanding dialogues. 
When compared to the fully-trained baselines, our proposed descriptor-based methods achieve superior performance even in zero-shot settings, establishing a new state-of-the-art performance achieved by a single model.

\begin{table}[t!]
\centering
\resizebox{0.48\textwidth}{!}{
\begin{tabular}{|l|ccc|c|}
\hline
\textbf{Ensemble} & \textbf{R@1} & \textbf{R@5} & \textbf{R@10}  & \bf Avg\\
\hline\hline
S + G & 47.32 & 69.62 & 77.63 & 64.86\\
S + Q & 47.78 & 68.81 & 77.61 & 64.73\\ 
G + Q & 47.44 & 68.90 & 77.15 & 64.50\\ 
S + G + Q & 48.79 & 70.01 & 78.44 & 65.75\\
S + G + Q + C & \textbf{48.84} & \textbf{70.20} & \textbf{78.74} & \bf 65.93\\ 
\hline
\end{tabular}}
\vspace{-1mm}
\caption{\label{tab: ensemble}
Ensemble results of fully-trained retrievers with LLaMA-2-13B-Chat as the LLM (\%). (S: Summary; G: Guessing; Q: Queries; C: Caption).}
\vspace{-3mm}
\end{table}

\begin{table*}[ht]
\centering
\resizebox{0.78\textwidth}{!}{
\begin{tabular}{|l|l|ccc|c|}
\hline
\textbf{Method} & \textbf{ Score } & \textbf{R@1} & \textbf{R@5} & \textbf{R@10} & \textbf{Avg} \\
\hline\hline
\multirow{2}{*}{Desc - Summary}  
 & Scene-Aligned (Text-Only) & 35.07 & 49.37 & 57.66 & 47.37 \\
 & Vision-Aligned (Multimodal) & 29.37 & 53.18 & 62.49 & 48.35 \\
\hline
\multirow{2}{*}{Desc - Guessing} 
 & Scene-Aligned (Text-Only) & \textbf{35.82} & 50.58 & 58.30 & 48.23 \\
 & Vision-Aligned (Multimodal) & 28.41 & 53.78 & 63.90 & 48.70 \\
\hline
\multirow{2}{*}
{Desc - Queries}  
& Scene-Aligned (Text-Only) & 35.53 & 50.64 & 58.68 & 48.28 \\
& Vision-Aligned (Multimodal) & 29.16 & \textbf{54.28} & \textbf{64.17} & \textbf{49.20} \\
\hline
\end{tabular}}
\vspace{-2mm}
\caption{\label{tab: ablation modality}
The results of the model trained using either scene-aligned or vision-aligned scores.
}
\vspace{-2mm}
\end{table*}

\subsection{Ensemble}
We further conduct experiments on ensemble learning using all descriptor-based results based on the validation set.
The results in Table \ref{tab: ensemble} demonstrate that ensemble learning consistently improves performance. 
Even in cases where the caption model performs poorly in a fully-trained setting, ensemble learning benefits other models.
These findings highlight the efficacy of combining various types of descriptors, leading to the best overall performance and establishing a new state-of-the-art for PhotoChat. 
This suggests that the generated descriptors focus on diverse patterns that can complement each other and enhance scores.


\section{Analysis}

\subsection{Effectiveness of Two Alignment Scores}

Our proposed method incorporates two scores: scene-aligned (text-only) and vision-aligned (multimodal) scores. 
We conduct an ablation study to assess the impact of each score. 
Table \ref{tab: ablation modality} presents the experimental results. 
The results show that models trained solely on the scene-aligned score (text-only) perform better in terms of $R@1$, whereas models trained on the vision-aligned score (multimodal) perform better for $R@5$ and $R@10$.

\begin{figure}[t]
\centering
\vspace{-5mm}
\includegraphics[width=.45\textwidth]{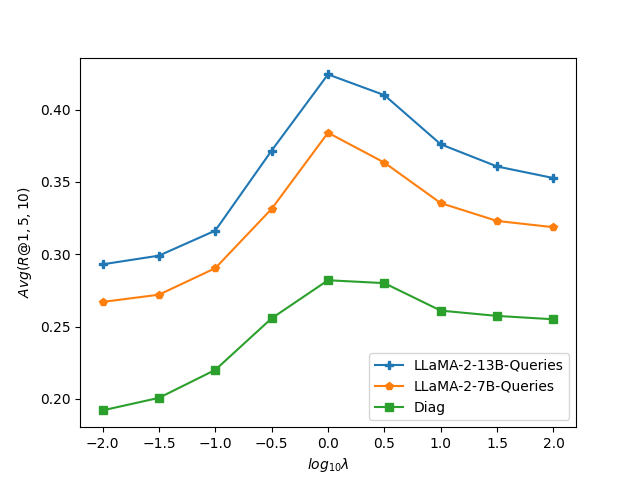}
\vspace{-2mm}
\caption{
Results of different $\lambda$ in zero-shot scenarios. A smaller $\lambda$ indicates greater reliance on the scene-aligned score, while a larger $\lambda$ indicates greater reliance on the vision-aligned score.
}
\label{img:lambda}
\vspace{-2mm}
\end{figure}

In addition, we conducted experiments with different weighting parameter $\lambda$ in (\ref{eq:retrieval score}). Figure \ref{img:lambda} illustrates the results obtained for various $\lambda$ values in the zero-shot scenarios. It can be observed that the best performance is achieved when $\lambda$ approaches 1. This indicates that excessive bias towards either the scene-aligned score or the vision-aligned score leads to a decrease in performance. Interestingly, our findings reveal that the optimal zero-shot performance occurs at $\lambda$ = 1.2.

\subsection{Object Detection Sensitivity}

Given our experimental assumption of having available object lists for the images, a degradation in system performance may occur when object detection is poor.
Consequently, we explored the sensitivity of error propagation in our proposed method by simulating various error rates in detected objects.
Specifically, we randomly removed objects from the gold object list or substituted them with others from the training set to simulate scenarios of ``missing'' or ``incorrect'' objects.
As shown in Table \ref{tab:object accuracy}, it is evident that with increasing error rates, regardless of the error type, performance tends to degrade.
This suggests that our method is still influenced by error propagation.
Moreover, the impact of incorrect objects is particularly severe since they may mislead the model in image selection. 
Nevertheless, even with a 35\% error rate, our results outperform relying solely on the vision-aligned score, indicating a relative robustness to missing objects. Therefore, precision in object detection during object list extraction is more important for our proposed approach.

\begin{table}[t!]
\centering
\resizebox{0.48\textwidth}{!}{
\begin{tabular}{|cl|c c c | c|}
\hline
\bf Error Rate &  \bf Type & \textbf{R@1} & \textbf{R@5} & \textbf{R@10} & \bf Avg\\
\hline\hline
\hline
0\% & - & 26.40 & 45.80 & 55.10 & 42.43\\
\hline
 & missing & 24.75 & 43.85 & 53.45 & 40.68\\
15\% & incorrect & 22.04 & 39.99 & 49.01 & 37.01\\
 & both & 23.29 & 42.13 & 50.74 & 38.72\\
\hline
 & missing & 23.05 & 42.17 & 51.50 & 38.91\\
25\% & incorrect & 19.93 & \it 37.41 & \it 45.59 & \it 34.31\\
 & both & 21.82 & 39.92 & 48.42 & 36.72\\
 \hline
 & missing & 21.91 & 40.61 & 49.89 & 37.74\\
35\% & incorrect & \it 17.24 & \it 33.87 & \it 41.66 & \it 30.92\\
 & both & \it 19.25 & \it 36.98 & \it 45.33 & \it 33.85\\
\hline
100\% & missing & 19.80 & 38.50 & 46.60 & 34.97\\
\hline
\end{tabular}}
\vspace{-2mm}
\caption{\label{tab:object accuracy}
Sensitivity of diverse object error rates of LLaMA-2-13B-Chat in zero-shot settings. \textbf{Both} denotes the random removal or replacement of objects. 100\% missing means that we use only the vision-aligned score for retrieval. The italicized fonts indicate scores lower than vision-aligned only results.
}
\vspace{-2mm}
\end{table}

\subsection{Visually-Focused Query Impact}

To assess the influence of different visually-focused queries on our results, we conducted experiments by systematically removing individual queries from the original query set (see Table \ref{tab:ablation - queries}). Remarkably, the query concerning the \emph{main subject} emerges as the most pivotal feature for bridging the dialogue context and the target image, as its removal leads to a significant decrease in scores. 
Following in descending order of impact are the queries related to \emph{foreground objects}, \emph{events}, \emph{background scene}, and \emph{materials and attributes}.


\begin{table}[t]
\centering
\resizebox{\columnwidth}{!}{
\begin{tabular}{|l|ccc|c|}
\hline
\bf Method               & \bf R@1    & \bf R@5    & \bf R@10   & \bf Avg \\ \hline\hline
Original               & 44.00 & 64.78 & \textbf{73.95} & 60.91\\ 
- main subject       & 28.80 & 49.16 & 58.41 & 45.56          \\
- foreground objects & 40.44 & 61.62 & 70.49 & 57.52          \\
- background scene   & 43.65 & 64.05 & 72.88 & 60.19\\
- events             & 42.91 & 64.00 & 72.78 & 59.90          \\
- materials \& attributes    & 43.22 & 64.60 & 73.59 & 60.47          \\
+ atmosphere or mood & 43.67 & 64.89 & 73.85 & 60.80          \\
+ lighting           & \textbf{44.13}	& \textbf{64.95} & 73.93 & \textbf{61.00}          \\ 
\hline
\end{tabular}}
\vspace{-2mm}
\caption{The results of different queries. All additions and removals are based on the original query set.}
\label{tab:ablation - queries}
\vspace{-2mm}
\end{table}

Additionally, we augmented the original query set with two common photo features: \emph{atmosphere or mood} and \emph{lighting}.
\begin{compactitem}
    \item \textbf{Atmosphere or mood:} Research by \citet{sun-etal-2022-multimodal} suggests that photographs convey not only object details but also emotions and ambiance, evoking feelings like happiness, boredom, and coziness.
    \item \textbf{Lighting:} As highlighted by \citet{hunter2021light}, the presence and quality of light are fundamental in photographic composition. Our investigation explores whether an LLM can accurately predict lighting conditions within a photo solely based on dialogue context and whether this predictive information enhances text-image retrieval capabilities.
\end{compactitem}
Results indicate that \emph{atmosphere or mood} improves performance at $R@5$, while \emph{lighting} enhances results at both $R@1$ and $R@5$ compared to the original set. This suggests that these abstract and challenging-to-predict queries have varying impacts on performance.

\begin{table}[t]
\centering
\begin{tabular}{|l|c c |}
\hline
\bf Method & \bf VisDial & \bf MMDialog\\
\hline\hline
\citet{koh2023grounding}  &  40.57 & -\\
\hline
Desc - Diag & 22.53 &  32.47\\
Desc - Summary &  35.77 & 19.03\\
Desc - Guessing &  \bf 49.29  & \underline{36.93}\\
Desc - Queries & \underline{44.45} & 36.40\\
Desc - Queries (all) &  43.80 & \bf 38.27\\
\hline
\end{tabular}
\vspace{-2mm}
\caption{\label{tab:generalization}
Average zero-shot performance on VisDial and MMDialog (\%). 
}
\vspace{-3mm}
\end{table}

\subsection{Generalization to Other Datasets}

To explore the generalizability of our approach, we conducted experiments on other benchmark datasets, VisDial~\cite{das2017visual} and MMDialog~\cite{feng-etal-2023-mmdialog}, with identical settings. 
\begin{compactitem}
    \item VisDial \cite{das2017visual}: Each instance in VisDial comprises a single image from the COCO dataset \cite{lin2014microsoft} and a dialogue context consisting of question-answer pairs. During testing, the model's task is to retrieve the correct image from a pool of 2,064 images in the testing set.
    \item MMDialog \cite{feng-etal-2023-mmdialog}: This dataset comprises multimodal open-domain dialogues gathered from a global social media platform. Instances in the testing set are linked with 999 corresponding negative images. MMDialog uniquely allows dialogue context to contain both text and images. In our experiments, we employed BLIP2 \cite{li2023blip} for image captioning within the dialogue context. These resulting captions replaced the images, enabling textual dialogue context input into the LLM. Our experiments were conducted on a random sample of 1,000 instances from the testing set.
\end{compactitem}
The average performance of LLaMA-2-13B-Chat in zero-shot scenarios across different datasets is summarized in Table \ref{tab:generalization}, with detailed scores available in Appendix~\ref{sec:other_data}. 
Our findings indicate that our proposed approach consistently outperforms {\bf Desc - Diag} baseline and even recent work~\cite{koh2023grounding}, showcasing its robust generalization capability across diverse datasets.

\subsection{Qualitative Study}

\begin{table*}[t]\centering
\small
\begin{tabular}{p{0.55\textwidth}p{0.38\textwidth}}\toprule
\textbf{Dialogue Context} & \textbf{Ground-Truth Photo}\\
\textbf{B:} whats up &\multirow{11}{*}{\includegraphics[width=0.23\textwidth,height=0.23\textwidth]{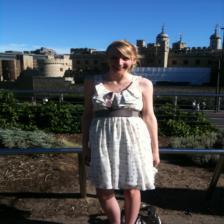}} \\
\textbf{A:} Hanging out with my student Maren & \\
\textbf{B:} Oh thats cool & \\
\textbf{B:} how was it ? & \\
\textbf{A:} Great. We are sightseeing castles & \\
\textbf{B:} That sounds really cool actually & \\
\textbf{B:} took any pictures of it? & \\
\textbf{A:} Yes & \\
\textbf{B:} Show me it I wanna see & \\
\textbf{A:} sure & \\
\textbf{A:} \colorbox{yellow}{(share a photo)} & \\
\end{tabular} \\
\begin{tabular}{p{0.31\textwidth}p{0.31\textwidth}p{0.31\textwidth}}\toprule
\multicolumn{3}{c}{LLaMA-2-7B-Chat} \\\midrule
\textbf{Desc - Queries} &\textbf{Desc - Summary} &\textbf{Desc - Guessing} \\
The main subject of the photo is a picture of a castle.
The prominent objects in the foreground of the photo is Maren and \textbf{A}.
The background scene of the photo is a \textcolor{red}{beautiful landscape with rolling hills} and a blue sky.
The events of the photo is sightseeing.
The materials and attributes of the photo is a \textcolor{red}{camera and a smartphone} & \textcolor{orange}{\textbf{A} is a teacher who is spending time with their student Maren.} They are sightseeing castles and took pictures of them. \textcolor{orange}{\textbf{A} is willing to show the pictures to \textbf{B}.} &It is likely that \textbf{A} shared a photograph of a castle that they and Maren visited during their sightseeing trip. The photograph may show the exterior or interior of the castle, or a specific feature or detail of the castle that caught \textbf{A}'s attention. 
\\
\end{tabular} \\
\begin{tabular}{p{0.31\textwidth}p{0.31\textwidth}p{0.31\textwidth}}\toprule
\multicolumn{3}{c}{LLaMA-2-13B-Chat} \\\midrule
\textbf{Desc - Queries} &\textbf{Desc - Summary} &\textbf{Desc - Guessing} \\
The main subject of the photo is Maren.
The prominent objects in the foreground of the photo are castles.
The background scene of the photo is countryside.
The events of the photo are sightseeing.
The materials and attributes of the photo are medieval architecture. & \textcolor{orange}{\textbf{A} is a teacher or mentor figure, as they are referred to as "my student Maren"} \textbf{A} and Maren are currently sightseeing castles together \textbf{A} has taken pictures of the castles they have visited \textcolor{orange}{\textbf{A} is willing to share these pictures with \textbf{B}} &\textbf{A} shared a photograph of a castle that they visited with their student Maren. The photograph is likely to be a scenic view of the castle, showcasing its architecture and surroundings. The photograph may include Maren in the frame, possibly posing in front of the castle or looking out at the view. 
\\
\bottomrule
\end{tabular}
\vspace{-2mm}
\caption{\textcolor{red}{Red} text represents incorrect information. \textcolor{orange}{Orange} text represents redundant sentences that do not aid in the text-to-image retrieval task.
}\label{tab: case-study}
\vspace{-2mm}
\end{table*}

To further explore the quality of our generated descriptors, we conducted a qualitative analysis across various models and descriptors. We present an example of a testing dialogue using different LLMs and descriptor generation methods in Table \ref{tab: case-study}.
A comparison between LLaMA-2-7B-Chat and LLaMA-2-13B-Chat in the \textbf{Desc - Queries} domain revealed that LLaMA-2-7B-Chat often delivers inaccurate or unrelated answers (highlighted in red).
In contrast, LLaMA-2-13B-Chat tends to produce more precise answers and demonstrates an ability to infer information not explicitly present in the dialogue context (e.g., medieval architecture).
This discrepancy is likely attributable to differences in reasoning capabilities between the two models.

Both \textbf{Desc - Summary} and \textbf{Desc - Guessing} can accurately describe the features of the photos. However, \textbf{Desc - Summary} occasionally incorporates additional details not directly related to the photos, such as ``\textit{A is willing to share these pictures with B},'' likely because we instructed the LLM to summarize information about the users sharing the photo rather than solely focusing on the photo-related information.
Our analysis suggests that our visually-focused approaches are particularly effective for this task, especially when coupled with the powerful reasoning capabilities of LLMs.

\section{Conclusion}

This paper presents a novel approach to enable multimodal dialogue systems for seamless photo sharing. By harnessing the reasoning capabilities of LLMs, our method generates precise visual cues from ongoing dialogue contexts. Addressing challenges encountered in prior methods, such as accurately understanding extensive dialogue contexts and handling input length constraints, our approach exhibits clear superiority in experimental results. Additionally, our comprehensive ablation study confirms the effectiveness of text-only visual descriptors, indicating a promising pathway for bridging intricate dialogues and images through deep dialogue understanding via LLMs. This work not only advances photo sharing within dialogues but also lays groundwork for future sophisticated multimodal dialogue systems.

\section*{Limitations}


Our method assumes the availability of object detection capabilities during pre-processing to extract object lists associated with the images. 
This reliance on object detectors may limit the method's applicability in scenarios where object detection is challenging or unavailable, potentially affecting its performance.

Lastly, our method assumes that the shared images align with the given dialogue context. In cases where users share images that are intentionally misleading, unrelated to the conversation, or dependent on users' personal information, our method may struggle to retrieve appropriate images, leading to potential accuracy issues in such scenarios.

\section*{Acknowledgements}
We thank the reviewers for their insightful comments. 
This work was financially supported by the National Science and Technology Council (NSTC) in Taiwan, under Grants 111-2222-E-002-013-MY3, 111-2628-E-002-016, and 112-2223-E002-012-MY5.
We thank to National Center for High-performance Computing (NCHC) of National Applied Research Laboratories (NARLabs) in Taiwan for providing computational and storage resources.

\bibliography{anthology,custom}

\appendix

\section{Prompts}
\label{appendix:prompts}

The designed prompts for all descriptor-based approaches are shown as follows.

\subsection{Desc - Summary}

\begin{tcolorbox}[width=\columnwidth,colback=white]
\small
\begin{verbatim}
Please read the following dialogue context: 
<dialogue_context>

Based on the dialogue context, please 
summarize the information of speaker A.

Answers: 
\end{verbatim}
\end{tcolorbox}

\subsection{Desc - Guessing}
\begin{tcolorbox}[width=\columnwidth,colback=white]
\small
\begin{verbatim}
Please read the following dialogue context: 
<dialogue_context>

Based on the dialogue context, please describe
the photograph shared by speaker A. 

Answers: 
\end{verbatim}
\end{tcolorbox}

\begin{table*}[t]
\centering
\begin{tabular}{|l|c c c c|c c c c|}
\hline
\multirow{2}{*}{\bf Method} &  \multicolumn{4}{c|}{\bf VisDial} & \multicolumn{4}{c|}{\bf MMDialog}\\
 &  \textbf{R@1} & \textbf{R@5} & \textbf{R@10} & \bf Avg & \textbf{R@1} & \textbf{R@5} & \textbf{R@10} & \bf Avg\\
\hline\hline
FROMAGe~\cite{koh2023grounding}  & 20.8~ & 44.9~ & 56.0~ & 40.57 & - & - & - & - \\
\hline
Desc - Diag & 11.00 & 24.03 & 32.56 & 22.53 & 22.10 & 34.30 & 41.00 & 32.47\\
Desc - Summary & 19.43 & 39.05 & 48.84 & 35.77 & 14.60 & 20.00 & 22.50 & 19.03\\
Desc - Guessing &  \bf 29.46 & \bf 53.73 & \bf 64.68 & \bf 49.29 & \bf 31.30 & 37.30 & \underline{42.20} & \underline{36.93}\\
Desc - Queries &  \underline{24.52} & \underline{49.13} & 59.69 & \underline{44.45} & 30.90 & \underline{37.50} & 40.80 & 36.40\\
Desc - Queries (all) & 24.3 & 47.34 & \underline{59.74} & 43.80 & \underline{31.10} & \bf 40.60 & \bf 43.10 & \bf 38.27\\
\hline
\end{tabular}
\vspace{-2mm}
\caption{\label{tab:other-result}
Retrieval performance for zero-shot on VisDial and MMDialog datasets (\%). \textbf{Desc-Queries (all)} include the original five queries along with actions, atmosphere, or mood, and lighting queries
We employ the LLaMA-2-13B-Chat with greedy decoding to ensure the correct format and reasoning capability.
}
\vspace{-2mm}
\end{table*}

\subsection{Desc - Queries}

\begin{tcolorbox}[width=\columnwidth,colback=white]
\small
\begin{verbatim}
Please read the following dialogue context: 
<dialogue_context>

Based on the dialogue context, please describe
the photograph shared by speaker A.
List the answer in JSON format. 
- main subject: {simply list the answer by ','} 
- prominent objects in the foreground: {simply 
list the answer by ','} 
- background scene: {one background scene} 
- events: {simply list the answer by ','} 
- materials and attributes: {simply list the 
answer by ','} 

Answers: 
\end{verbatim}
\end{tcolorbox}



\section{Detailed Results for Other Datasets}
\label{sec:other_data}

Table \ref{tab:other-result} presents detailed scores on the VisDial and MMDialog datasets using LLaMA-2-13B-Chat in zero-shot scenarios. 
It is worth noting that we sampled 1,000 instances from the MMDialog testing data for our experiments, limiting direct comparison with previous work due to the discrepancy of testing sets.





\end{document}